# Alternative Technique to Asymmetry Analysis-Based Overlapping for Foot Ulcer Examination: Scalable Scanning

Naima Kaabouch[1]*, Wen-Chen Hu[2] and Yi Chen[1]

[1]Electrical Engineering Department, University of North Dakota, ND 58202, USA
[2]Computer Science Department, University of North Dakota, ND 58202, USA

**Abstract**

Asymmetry analysis based on the overlapping of thermal images proved able to detect inflammation and, predict foot ulceration. This technique involves three main steps: segmentation, geometric transformation, and overlapping. However, the overlapping technique, which consists of subtracting the intensity levels of the right foot from those of the left foot, can also detect false abnormal areas if the projections of the left and right feet are not the same. In this paper, we present an alternative technique to asymmetry analysis-based overlapping. The proposed technique, scalable scanning, allows for an effective comparison even if the shapes and sizes of the feet projections appear differently in the image. The tested results show that asymmetry analysis- based scalable scanning provides fewer false abnormal areas than does asymmetry analysis -based overlapping.

## Introduction

Foot ulceration is an important precursor to amputation, having been identified as a component in 84% of lower extremity amputations [1,2]. Obesity and Chronic conditions associated with an aging U.S. population - such as diabetes, neuropathy, circulatory insufficiency, or a combination of these pathologies - mirror the current increased incidence of foot ulcers. The heel is the second most common site of pressure ulcers, producing 28% of all reported pressure ulcers [3]. These ulcers are among the most difficult to heal. In 1995 alone, lower-extremity ulcers cost Medicare $1.5 billion [4]. Patients with foot ulcers also can suffer from secondary conditions, including severe pain, immobility, increased infection risks, embarrassment and worry, and a dramatic impact on daily quality of life.

One of the most common mechanisms that produce foot ulceration involves a cumulative effect of unrecognized repetitive trauma at pressure points on the sole of the foot over the course of several days [5-7]. Areas that are likely to ulcerate have also been associated with increased local skin temperatures due to inflammation and enzymatic autolysis of tissue [8-10]. Identifying precise areas of injury by the presence of inflammation can allow patients or healthcare providers to take early action to decrease the inflammation before a wound or ulcer actually develops. Such inflammation is mainly characterized by five signs: Redness, pain, swelling, loss of function, and heated tissue. Some of these signs are difficult to assess objectively. In a neuropathic extremity, pain and disturbance of function may be absent because of neuropathy, and thus, these signs are poor indicators of oncoming inflammation [11]. In addition, swelling and redness are difficult to grade precisely even for experienced clinicians.

Conventional noninvasive methods to assess skin, including visual inspection and palpation, can be valuable diagnostic methods, but usually they do not detect complete enough changes in skin integrity until skin breakdown has actually occurred. However, temperature measurements can provide quantitative information that can be more precisely predictive of impending ulceration [8,12-14].

In this research work, thermal imaging is used to monitor the temperature distribution of foot skin. However, there is no standard distribution for the skin-surface temperature of a healthy foot because that temperature can be affected by many factors, such as ambient and internal thermal conditions, age, sex, weight, etc. One way to eliminate this variability is to compare the thermal skin distributions of both feet on the same subject [10,11]. This comparison, called asymmetry/symmetry analysis, has been widely used by researchers and clinicians to identify pathological conditions in the brain, breast, and other body parts that present similar symmetric characteristics.

In a previous research [15-18], asymmetry analysis was combined with a genetic algorithm to detect inflammation and, hence, predict ulcers before they can develop. The methodology involves three steps:

- Segmentation-isolate the feet and remove as much noise as possible,
- Geometric transformation-adjust the left and the right foot so the two are in the same position in the image,
- Asymmetry analysis-subtract the intensity level of each pixel in the left foot from the intensity level for the symmetric pixel of the right foot to detect abnormal areas. In each foot, an abnormal area is detected if the intensity level (temperature) of that area is higher than a specific designated threshold.

Although this technique shows high efficiency in identifying inflammation when the feet present as the same size and shape in an image, it tends to detect false abnormal areas when the feet are not the same size or shape in the image. In this paper, an alternative technique to asymmetry analysis- based overlapping is offered. The proposed technique, asymmetry analysis- based scalable scanning, allows for a more reliable comparison even if the shapes and sizes of the two foot projections are different.

## Methodology

A high resolution thermal camera, FLIR A320 (FLIR Systems Inc., Boston, USA) with a thermal sensitivity less than 0.05°C, was used to record the temperature of foot skin distributions 5 minutes after the patients have removed their socks and shoes. Successive images of the patient's feet are taken during 15 minutes. However, in this paper only the first image of the patient's feet is analyzed. The resulting thermal image is then analyzed using an approach, implemented using MATLAB as a platform involving the following steps:

1. Segment the image
2. Divide the image into separate images, each containing the left and the right foot
3. Find the centroid of each foot and the furthest point from the centroid to the heel edge in each image
4. Initialize
- Put edge points into a queue and index each point
- Determine the angle step and the number of lines
5. Calculate the angle of $i_{th}$ line
6. Find the intersection point of the $i_{th}$ line with the foot edge
7. Find the straight line running from the centroid to the intersection point, using the Bresenham algorithm
8. Determine the comparison points along the $i_{th}$ line
9. Scan the whole foot, and compare the points on every line in the left foot to the points of the corresponding line in the right foot
10. Identify abnormalities, if any, in each foot.

In the first step, a genetic algorithm was used to extract the feet from the image background and remove as much noise as possible. Genetic algorithms are optimization algorithms, based on biological mechanics of natural selection, such as chromosome, population size, cross rate, mutation rate, and maximum generation [16,19-20].

The main steps of the genetic algorithm are summarized below:

1. Assign the length of chromosome, population size, cross rate, mutation rate, and maximum generation.
2. Initialize population of the first generation, with each individual being a random eight bits binary string, representing a specific intensity level.
3. Evaluate the fitness of the whole population.

The fitness is evaluated by

$$Fitness(x) = Num_f \cdot Num_b \cdot (M_f - M_b)^2 \qquad (1)$$

Where $Num_f$ and $Num_b$ are the numbers of foreground and background pixels, respectively.

$M_f$, the mean intensity of foreground pixels, is given by

$$M_f = \frac{I_f}{Num_f} \qquad (2)$$

Where $I_f$ is the sum of intensities of foreground pixels.

$M_b$ is the mean intensity of background pixels given by

$$M_b = \frac{I_b}{Num_b} \qquad (3)$$

Where $I_b$ is the sum of intensities of background pixels.

4. Generate the next population by performing selection, crossover and mutation operations.
5. Go to step 3 if the desired number of generations is not reached.
6. Reduce the cross rate and mutation rate after half of the desired generation number is reached.
7. Segment the image using the optimal threshold level when the desired number of generation is reached.

The best threshold is determined by the following equation:

$$Fitness(x^*) = \max\{Fitness(x)\} \qquad (4)$$

Where $x$ represents a population

The output image of the genetic algorithm is then divided in two images, containing the left and the right foot, respectively. After this process, the centroid and the farthest point from the centroid to the heel edge are identified in each image and then used as feature points for a clear reference line.

In the fifth step, the resolution of the scanning is defined. This resolution depends on two parameters: 1) the number of radial lines and 2) the distance between two comparison points. The first parameter, the number of radial lines, represents how many radial lines are compared in a foot with the same corresponding radial lines in the other foot. For each foot, the location of each line is determined by the angle step given as:

$$anglestep = 2*pi/lines \qquad (5)$$

The number of lines is set in such a way that the anglestep will not exceed five degrees to scan the smallest area possible. The second parameter, the distance between two comparison points, serves as the scale adopted for our comparison. It represents the number of pixels between the two comparison points in the same radial line. A standard distance is used on the left foot, and the distance is scaled line by line on the right foot to guarantee the same number of comparison points. The edge points of each foot are sorted counter-clockwise with the heel point as the initial point. The radial lines are constructed counter-clockwise with each edge point representing the limit of each line.

As shown in Figure 1, a reference line is set up between the two feature points, the centroid and the heel point. The second line is located at one angle step from the reference line, the third line is located at two angle steps from the reference line, and so on. To limit the scanning within the feet, the edge points of the feet are used as the limits of the radial lines. Figure 1 shows examples of radial line point limits for a specific angle step.

While it is easy to find the coordinates of points within a continuous line, it is difficult to determine discrete points (pixels) along a line in computer graphics. In our approach, a well-known algorithm called the Bresenham line drawing was used. This algorithm computes the discrete best-fit line from (0, 0) to (X, Y), where the point (X, Y) lies in the NE half-quadrant, i.e., 0 < Y =<X. The best-fit line is one that does not deviate more than half a pixel away from the real line. For efficiency, the algorithm computes the pixel values (x, y) of this best-fit line using only linear operations. By comparing the centroid and the foot edges point with the matrix of the Bresenham points, the pixels belonging to each radial line can be determined. As shown in Figure 1, the Bresenham points for each radial line are defined between the centroid, and the intersection edge point is clipped for a specific angle step.

In the initialization section, the left foot always uses a standard distance which is assigned manually. The distance of the right foot varies line by line as follows:

$$stepR(i) = stepL * Ratio(i) \quad (6)$$

Here, Ratio(i) is given by

$$Ratio(i) = len\_Right(i)/len\_Left(i) \quad (7)$$

The len_Right(i) and the len_Left(i) represent the lengths of the corresponding segments of the left foot and right foot in the image, respectively.

After the grid matrices of the left and the right feet are generated in a corresponding fashion, any abnormalities are identified by comparing each point in the radial line to its corresponding point in the line of the other foot. This comparison is made from the initial line to the assigned number of lines and from the centroid to the edge point along the line. A threshold is assigned manually to determine which of the two compared points is abnormal. If the difference in the intensities of the left foot and the right foot exceed this threshold, the point with the higher intensity is considered as an abnormal point and then marked out on the corresponding foot.

## Results and Discussion

To test the proposed algorithm, 140 thermal images were used. These images were divided into two sets. The first set included 80 images that corresponded to healthy feet with 40 images that represented feet with visibly different size projections and 40 images representing feet with the same size projections. The second set included 60 images corresponding to feet with visible thermal abnormalities, with 30 images having visibly different sizes of feet projections; the other 30 images had the same size of feet projections. Table 1 summarizes the images used for each set and category.

Examples of images from these sets are shown in Figures 2a, 2b. Figure 2a shows a thermal image of healthy feet, but with different feet projections in the image; Figure 2b shows an example of a thermal image in which the right foot contains a high temperature area, indicating the presence of inflammation. As one can see, these images have strong non-uniform backgrounds, mainly due to heat transfer from the legs and the feet.

Results after segmenting the images with the genetic algorithm, using Step 1 of the proposed algorithm, are shown in Figures 3a,3b, which correspond to the input images shown in Figures 2a,2b. As observed from the two images of Figures 3a,3b, the genetic algorithm efficiently segmented the feet from the background and removed much of the noise. The efficiency of this genetic algorithm was extensively studied in previous works [15,16].

Figure 4 shows the results after applying Steps 2 to 8 on the image of Figure 2b. The red spots in the images correspond to the grid matrices that correspond to the left and right foot, respectively. The spots give the locations of the points used for a comparison of the right and left feet for a specific angle step. Figure 5 offers the results after applying the proposed technique to the thermal image shown in Figure 2b. The red points in each foot represent those points that have higher intensity levels than those for the corresponding points in the other foot. As expected, an abnormal area was detected on the right foot.

In addition, other small abnormal temperature areas, especially those of the foot edges, were identified. These small areas, however, do not correspond to inflammations. Such false detection occurs

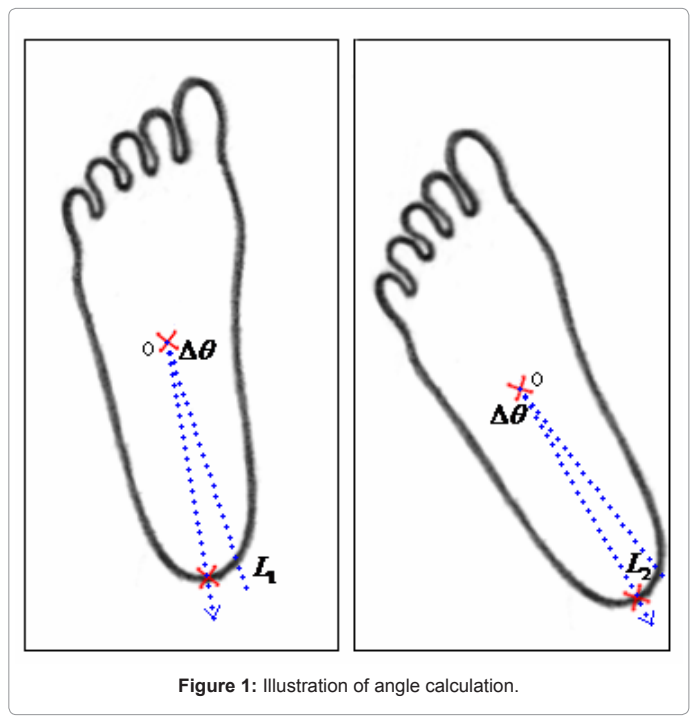

**Figure 1:** Illustration of angle calculation.

|  | Total Number of Images | Different Sizes | Same Size |
|---|---|---|---|
| Healthy feet | 80 | 40 | 40 |
| Feet with abnormalities | 60 | 30 | 30 |

**Table 1:** Number of images in each set used for the assessment.

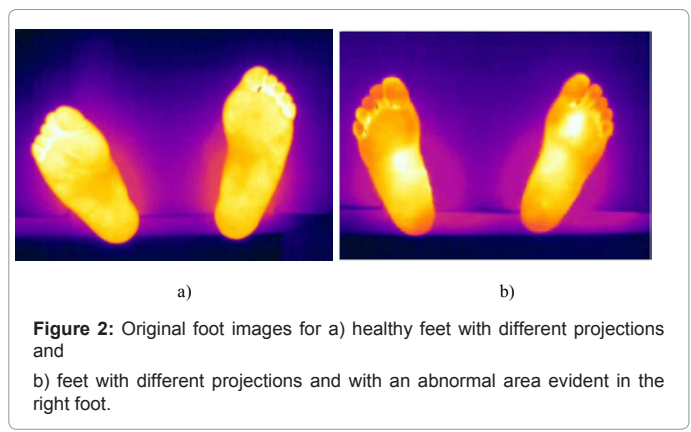

a)          b)

**Figure 2:** Original foot images for a) healthy feet with different projections and

b) feet with different projections and with an abnormal area evident in the right foot.

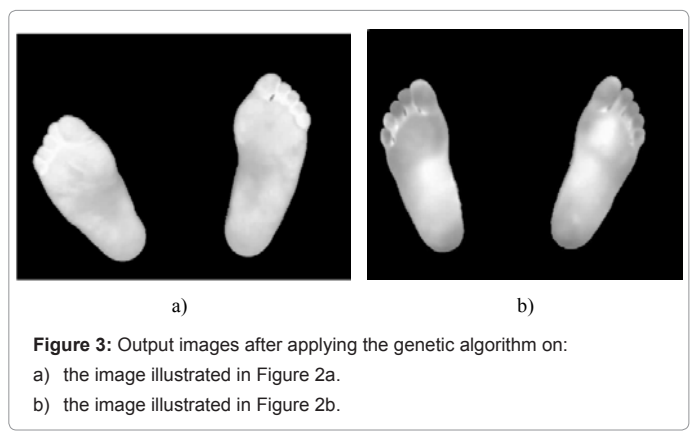

a)          b)

**Figure 3:** Output images after applying the genetic algorithm on:

a) the image illustrated in Figure 2a.

b) the image illustrated in Figure 2b.

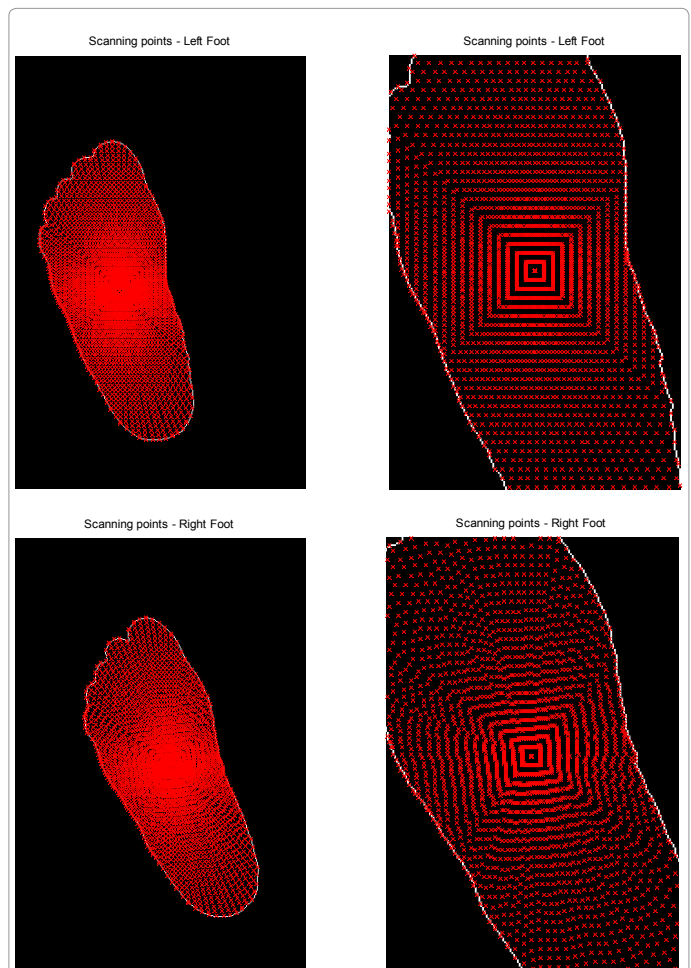

**Figure 4:** Location of the comparison points after applying steps 1 to 8 on the left and right feet images corresponding to the image of Figure 2b.

Figure 7 shows the results after applying the overlapping technique and the proposed technique, respectively, on Figure 2a, which contains healthy feet with different projections. One can see that the asymmetry-based overlapping technique indicates many false abnormal areas located on the edges, while the scanning technique offers a limited number of false abnormal points, located mainly on the foot edges. As mentioned earlier, such false detection is easily decreased by eliminating any comparison that involves foot edge points.

An assessment of the proposed technique utilizes a comparison using three criteria: the ability to detecting abnormal areas, the ability to analyze images with different feet projections without false detection, and the number of total false abnormal points. This latter number, however, depends on the image itself, i.e. the movement of the feet before recording the image. Table 2 provides a summary of this evaluation. One can see that while the overlapping technique works well only when the feet projections sizes and shapes are the same, the scalable scanning technique works well for all types of feet projections.

## Conclusion

In this research, we developed an alternative technique to using asymmetry analysis-based overlapping to investigate the thermal when the comparison threshold is set to an intensity equal to 5 and corresponding to less than 1°C, a difference not significant enough to distinguish between a normal and an abnormal area according to several studies, including those of Armstrong et al. [21] and Lavery et al. [22], a difference of 2.2° is more clinically indicative of impending ulceration, . Therefore any increase of the threshold should eliminate such false small areas without affecting the precision of the proposed technique. Figures 6 and 7 show the output images resulting from an increase in the threshold of the intensity values. As one can see, there is a decrease of false abnormal points on the foot edges. These false abnormal points will easily decrease when not involved with foot edge points in the comparison process.

To assess the performance of the proposed algorithm, we used the 140 images described in this section and compared the results of the proposed technique to those of asymmetry analysis- based overlapping. Figure 6 shows these results after applying both techniques on the image in Figure 2b. As one can see, both the overlapping and scalable scanning methods can identify abnormalities. However, the first technique detects false abnormal points on foot edges because the feet sizes are not perfectly equal, while the second technique detects fewer false abnormal points that correspond to the points located at the foot edges. This false detection can be minimized by eliminating any point comparison with the points located on the edges of the foot.

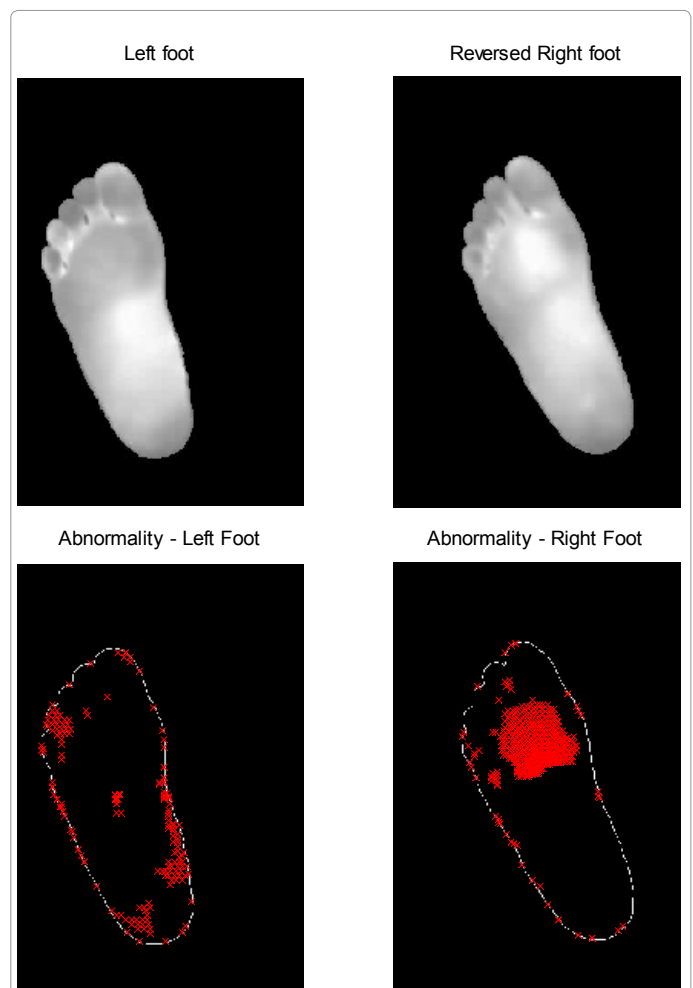

**Figure 5:** Red spots correspond to the identified abnormal areas in each foot after applying the scalable scanning to the original image illustrated in Figure 2b.

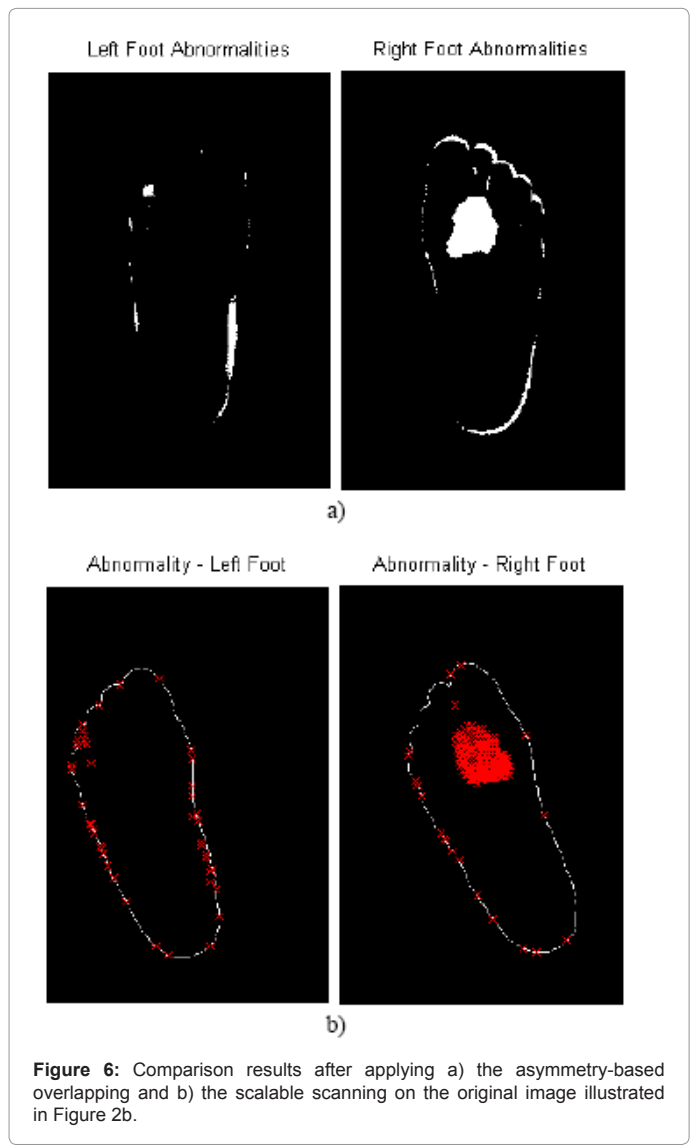

**Figure 6:** Comparison results after applying a) the asymmetry-based overlapping and b) the scalable scanning on the original image illustrated in Figure 2b.

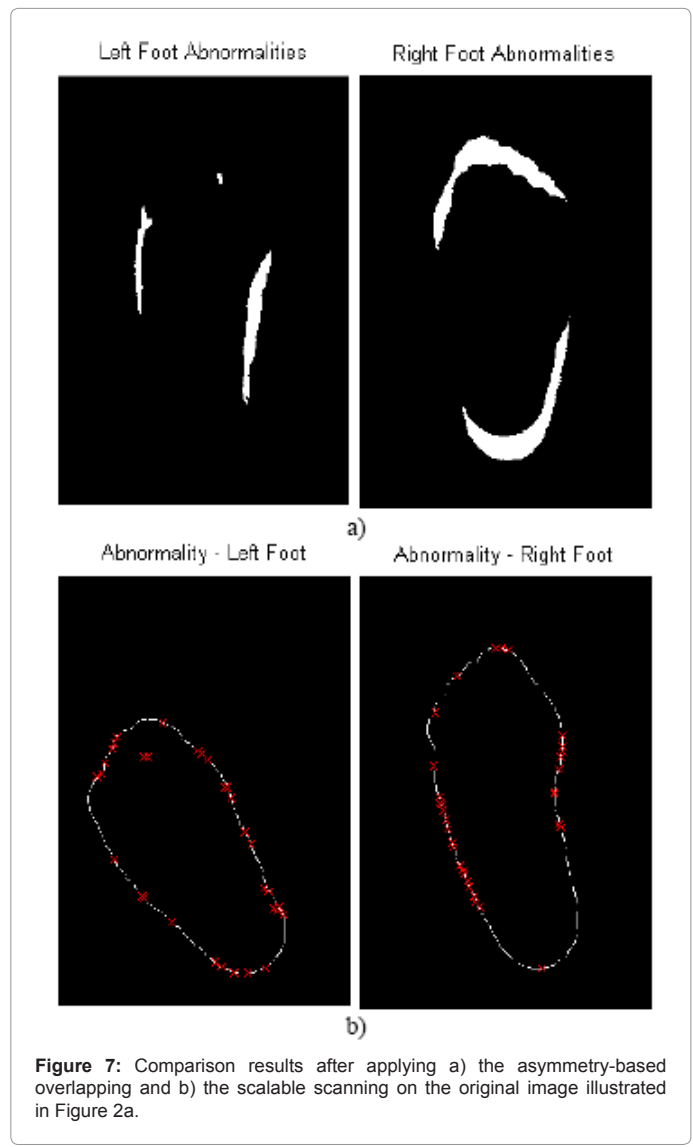

**Figure 7:** Comparison results after applying a) the asymmetry-based overlapping and b) the scalable scanning on the original image illustrated in Figure 2a.

|  | Total number of images without abnormal areas-- Healthy feet same sizes and shapes 40 | Total number of images without abnormal areas-- Healthy feet different sizes and shapes 40 | Total number of images with abnormal areas-- Feet of same sizes and shapes: 60 | Total number of images with abnormal areas-- Feet of different sizes and shapes 30 |
|---|---|---|---|---|
| Technique- | Number of images with false abnormal areas detected | Number of images with false abnormal areas | Number of images with abnormal areas detected | Number of false abnormal points |
| Asymmetry Based Overlapping | 0 | 40 | 60 | High |
| Proposed Technique – Scalable Scanning | 0 | 0 | 60 | Small |

**Table 2:** Results of the assessment.

distributions of feet to detect inflammation and predict foot ulcers. While, the overlapping technique works well when feet projection sizes are the same, it fails when feet projections are different. However, the scalable scanning technique works for all types of feet projections. The experimental results show that the proposed technique gives fewer false abnormal areas and hence is more reliable to use to identify inflammation. In the future, our research objectives will include 1) studying the decay rates of temperature distributions over time for successive images that correspond to the same feet, and 2) combining temperature and pressure distributions to predict foot ulceration.

### References

1. Pecoraro RE, Reiber GE, Burgess EM (1990) Pathways to diabetic limb amputation. Basis for prevention. Diabetes Care 13: 513-521.